# Generative AI in Agriculture: Creating Image Datasets Using DALL.E's Advanced Large Language Model Capabilities

**Ranjan Sapkota[*], Manoj Karkee[*]**

Cornell University, Department of Biological and Environmental Engineering, Ithaca, 14850, NY, USA.

## ABSTRACT

**The field of agricultural communication is evolving rapidly with the advent of generative artificial intelligence (AI), particularly image generation technologies. As these tools begin to influence how agricultural data is visualized and disseminated, the sector's diversity - spanning both technical and non-technical researchers, demands a rigorous foundational study to demystify the image generation process. This research investigated the role of artificial intelligence (AI), specifically the DALL.E model by OpenAI, in advancing data generation and visualization techniques in agriculture. DALL.E, an advanced AI image generator, works alongside ChatGPT's language processing to transform text descriptions and image clues into realistic visual representations of the content. The study used both approaches of image generation: text-to-image and image-to-image (variation). Six types of datasets depicting fruit crop environment were generated. These AI-generated images were then compared against ground truth images captured by sensors in real agricultural fields. The comparison was based on Peak Signal-to-Noise Ratio (PSNR) and Feature Similarity Index (FSIM) metrics. The image-to-image generation exhibited a 5.78% increase in average PSNR over text-to-image methods, signifying superior image clarity and quality. However, this method also resulted in a 10.23% decrease in average FSIM, indicating a diminished structural and textural similarity to the original images. Similar to these measures, human evaluation also showed that images generated using image-to-image-based method were more realistic compared to those generated with text-to-image approach. The results highlighted DALL.E's potential in generating realistic agricultural image datasets and thus accelerating the development and adoption of imaging-based precision agricultural solutions.**

**Keywords:** Text-to-image, Agriculture Communications, Image-to-image, Large Language Model (LLM), ChatGPT, OpenAI, Generative Artificial Intelligence (AI)

## Introduction

In recent years, synthetic images, generated by computer algorithms to resemble real-world entities, have been widely used in various sectors, such as healthcare(Chen et al. 2021), biomedicine(Barrera et al. 2023), fashion(Choi, Park, and Park 2022), architecture(Bermano et al. 2022), geospatial studies(Luo et al. 2022), automotive industry(Rio-Torto et al. 2021), and agriculture(Sapkota et al. 2022) due to their ability to provide realistic visual representations for observation, and analysis, and driving innovations(Man et al. 2022).

Traditional methods of creating these images include parametric techniques such as [9]Bezier Curves used by Chen et al. (A. Chen et al. 2021) to develop more accurate synthetic images of cell nuclei for biomedical applications. Similarly, Alberto et al. (Álvarez-Trejo et al. 2021) used Bezier Curves for designing complex mechanical structures using synthetic images. Another classical method is ray tracing, a technique to render realistic images by simulating light paths, improved upon by Ben et al. (Mildenhall et al. 2022) through Neural Radiance Fields for better low-light image reconstruction. Additionally, Physics-based Rendering (PBR) (Hodaň et al. 2019) , a method that mimics real-world light flow, has been used effectively for creating photorealistic images, as demonstrated by Hodan et al. (Hodaň et al. 2019) in their AI-based object detection research. These traditional image generation methods highlighted the evolving role of synthetic images in driving technological advancements (Zhao et al. 2022). However, these traditional models of synthetic image generation come with notable limitations. Parametric models, for instance, rely heavily on the accuracy of parameters and equations, making them less adaptable to complex or irregular shapes(Velikina et al. 2013) and environments that don't align with predefined mathematical structures (Araújo et al. 2018). The ray tracing technique faces challenges due to high computational intensity and time(Diolatzis, Philip, and Drettakis 2022). This technique also can be limiting in simulating complex lighting effects like indirect light and reflections (Zhang et al. 2022). PBR, on the other hand, has reduced flexibility and high computational demands(Eversberg et al. 2021).

Generative Adversarial Networks (GANs) present a promising alternative to generate synthetic images. These networks, consisting of a generator and a discriminator, efficiently produce realistic synthetic images(Wu, Xu, and Hall 2017). GANs provide greater flexibility than parametric and other traditional models by learning from high-dimensional data distributions, enabling them to generate more realistic images, even in complex scenarios [22]. This approach also addresses the computational challenges of ray tracing, as trained GANs can quickly generate new images (Zhang et al. 2022). Furthermore, GANs balance the realism-flexibility trade-off better than the PBR method, allowing detailed image generations without sacrificing quality(Matuszczyk et al. 2022) Their ability to create realistic images in complex environments enable them to be adoptable to wider fields and applications(Yu et al. 2018), making them a crucial tool in synthetic image generation.

In recent years, the application of Generative Adversarial Networks (GANs) in agriculture has gained increasing attention, particularly for tasks like disease detection and image augmentation, yielding promising results. For instance, Abbas et al.(Abbas et al. 2021) demonstrated the effectiveness of GANs, specifically using a Conditional GAN (C-GAN), to generate synthetic images of tomato plant leaves. This technique, combined with a DenseNet121 model and transfer learning, achieved a high accuracy of 99.5% in classifying tomato leaf diseases. It's noteworthy that their approach integrated both synthetic and actual images to enhance classification accuracy, suggesting a blend of novel and traditional methodologies. Furthermore, Lu et al. (Lu, Rustia, and Lin 2019) utilized GANs to create synthetic images of insect pests, thereby augmenting limited actual datasets (collected with sensors). This innovation significantly improved the performance of classifiers for insect pests, highlighting the utility of GANs in scenarios where actual data is scarce. Nazki et al. (Nazki et al. 2019) explored a different aspect of GANs by employing them for image-to-image translation in plant disease datasets, which facilitated more accurate disease classification. These studies indicate a trend towards using GANs not just for dataset augmentation - a role typically filled by conventional techniques like rotation and flipping - but also as a crucial tool in synthesizing and enhancing the quality of agricultural datasets. This shift marks a significant advancement in the application of AI in agriculture, opening new pathways for research and practical applications in the field.

Studies(Abbas et al. 2021; Liu et al. 2020) have shown that GANs effectively address the challenges of biological variability and the complexity of unstructured agricultural environments by successfully identifying and classifying pest and plant leaf diseases. GANs have been instrumental in several key areas of agricultural image processing. GANs enhance model efficiency by reducing the need for extensive data collection and labeling, particularly in diverse crop scenarios(De et al. 2022). For instance, Gomaa et al. (Gomaa and Abd El-Latif 2021) utilized a combination of Convolutional Neural Networks (CNN) and GANs for disease detection in tomato plants, highlighting the synergistic potential of combining traditional and generative models. Similarly, Madsen et al.(Madsen et al. 2019) applied Wasserstein auxiliary classifier generative adversarial networks (Wac-GAN) to model seedlings of nine different plants, showcasing the versatility of GANs in handling varied crop types. Zhu et al. (Zhu, He, and Zheng 2020) took a specialized approach with Conditional Deep Convolutional GANs (C-DCGAN) for orchid seedling vigor rating, emphasizing the precision capabilities of GANs. Further, studies like Hartley et al. (Hartley et al. 2021) with wheat for plant head detection using CycleGAN (Hartley and French 2021), and Bird et al. (Bird et al. 2022) focusing on lemon quality assessment using C-GAN illustrate the broad applicability of these networks across different crop environment.

Table 1 summarizes these recent efforts, showcasing how the integration of GANs in synthetic image generation is revolutionizing agricultural applications and contributing to the advancement of machine vision systems in agriculture.

**Table 1: Overview of GAN-based Synthetic Image Generation in Agriculture (2019-2024), Highlighting Image Generation Techniques, Crops, and Key Achievements.**

| Author Reference | Target Crop | Synthetic Image Generation Technique | Primary Objective |
|---|---|---|---|
| (Abbas et al. 2021) | Tomato plants | Conditional Generative Adversarial Network (C-GAN) | Disease detection |
| (Gomaa et al. 2021) | Tomato plants | Convolutional Neural Network (CNN) and GAN | Disease detection |
| (Madsen et al. 2019) | Nine different plant seedlings as:<br>1. Charlock<br>2. Cleavers<br>3. Common Chickweed<br>4. Fat Hen<br>5. Maize<br>6. Scentless Mayweed<br>7. Shepherd's Purse<br>8. Small-flowered Cranesbill<br>9. Sugar Beets | Wasserstein auxiliary classifier generative adversarial network (Wac-GAN) | Modeling plant seedlings |
| (Zhu et al. 2020) | Orchid seedlings | Conditional deep convolutional generative adversarial network (C-DCGAN) | Plant Vigor rating |
| (Hartley and French 2021) | Wheat | CycleGAN | Plant head detection |
| (Bird et al. 2022) | Lemons | C-GAN | Fruit quality assessment and defect classification |
| (Shete et al. 2020) | Maize plants | TasselGAN and deep convolutional generative adversarial networks (DCGAN) | Image generation of maize tassels against sky backgrounds |
| (Guo et al. 2021) | Jujubes | DCGAN | Quality grading |
| (Drees et al. 2021) | Arabidopsis thaliana and cauliflower plants | C-GAN (Pix2Pix) | Laboratory-grown and field-grown image generation |
| (Kierdorf et al. 2022) | Grapevine Berries | C-GAN/CDCGAN | Estimation of occluded fruits |
| (Olatunji et al. 2020) | Kiwifruit | C-GAN | Filling in missing fruit surface (Re-construction) |
| (Bellocchio et al. 2020) | Apple orchard | CycleGAN | Unseen fruits counting |
| (Fawakherji et al. 2021) | Sugar beet, sunflower | CGAN/CDCGAN | Crop/weed segmentation in precision farming |
| (Zeng et al. 2020) | Citrus | DCGAN | Disease severity detection |
| (Kim et al. 2021) | Blueberry leaves | DCGAN | Fruit tree disease classification |
| Tian et. al (Tian et al. 2019) | Apple canopy | CycleGAN | Disease detection |
| Cap et. al (Cap et al. 2020) | Cucumber leaves | CycleGAN | Plant disease diagnosis |

| (Maqsood et al. 2021) | Wheat | super-resolution generative adversarial networks (SR-GAN) | Wheat stripe(yellow) rust classification |
| --- | --- | --- | --- |
| (Bi and Hu 2020) | Grape, Orange, Potato, Squash, Tomato | Wasserstein generative adversarial network with gradient penalty (WGAN-GP) | Plant disease classification |
| (Zhao et al. 2021) | Apple, Corn, Grape, Potato, Tomato | DoubleGAN | Plant disease detection |
| (Nerkar and Talbar 2021) | Apple, corn, tomato, potato | Reinforced GAN | Leaf disease detection |

Building upon the foundational advancements introduced by Generative Adversarial Networks (GANs) in agricultural image processing, the DALL·E model by OpenAI (OpenAI, California, USA) represents a significant leap forward in the domain of AI-based image generation. Integrating the principles of GANs with innovative technologies such as Compact Language-Image Pretrained (CLIP) embeddings (Zhao et al. 2023), and Principal Component Analysis (PCA) for dimensionality reduction (Schuhmann et al. 2022), DALL·E transcends the capabilities of traditional image generation methods (Van Dis et al. 2023). One of the major breakthroughs with the DALL·E model is that it can convert textual descriptions into realistic images. Additionally, the model can generate a variation within an image representing similar environments. This capability of the model is achieved using text-conditional hierarchical image generation strategy (Schuhmann et al. 2022). Building on the foundation of ChatGPT developed by the same organization (OpenAI, California, USA), this model has been trained on an extensive variety and size of image-text pairs. Both models, stemming from the same OpenAI lineage, manifest exceptional competence in managing intricate, multi-dimensional tasks (McLean et al. 2023). For instance, while ChatGPT excels at generating contextually relevant textual responses, DALL·E emerges as a powerhouse in producing images that accurately represent the semantics of the input text(Liebrenz et al. 2023). Even though synthetic image generation has become easier and more accessible while providing more realistic images with OpenAI's DALL.E model (Link: https://chatgpt.com/g/g-2fkFE8rbu-dall-e), there is a need to thoroughly assess and evaluate its capability in representing field environments and its practicality in agricultural applications. To address this need, the following specific objectives were pursued in this study:

- To assess and evaluate the DALL·E model's proficiency in translating detailed textual prompts into accurate and realistic visual representations using text-to-image generation feature of the model for modern agriculture communications.
- To evaluate DALL·E model's ability to accurately transform an image prompt into generating realistic images of the similar environment using image variation feature of the model.

## Methods

### Data Collection and Compilation

In this study, the focus of image analysis was six distinct fruit crops datasets, as shown in Figure

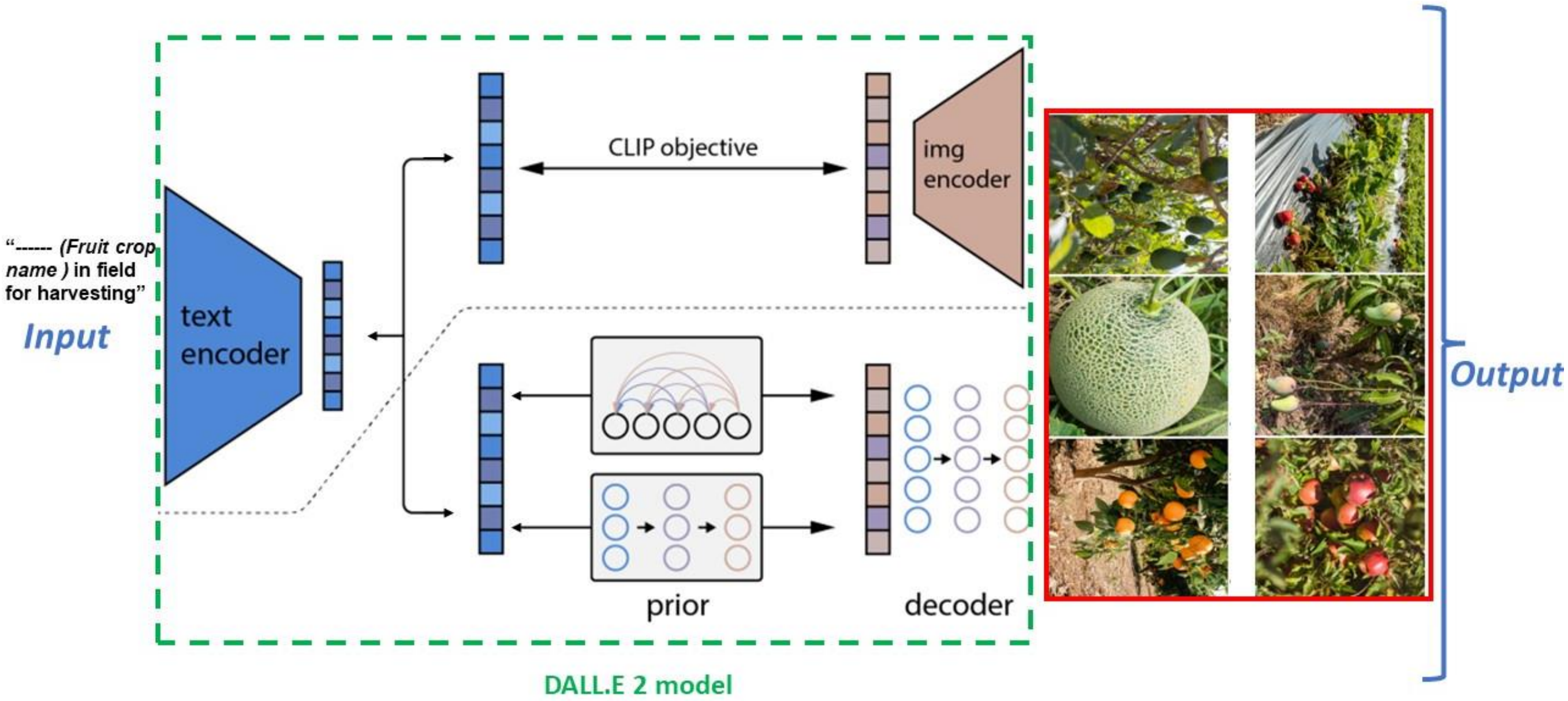

**Figure 1: Workflow showing process of utilizing DALL.E for dataset creation in this research focusing on fruit crops in agriculture**

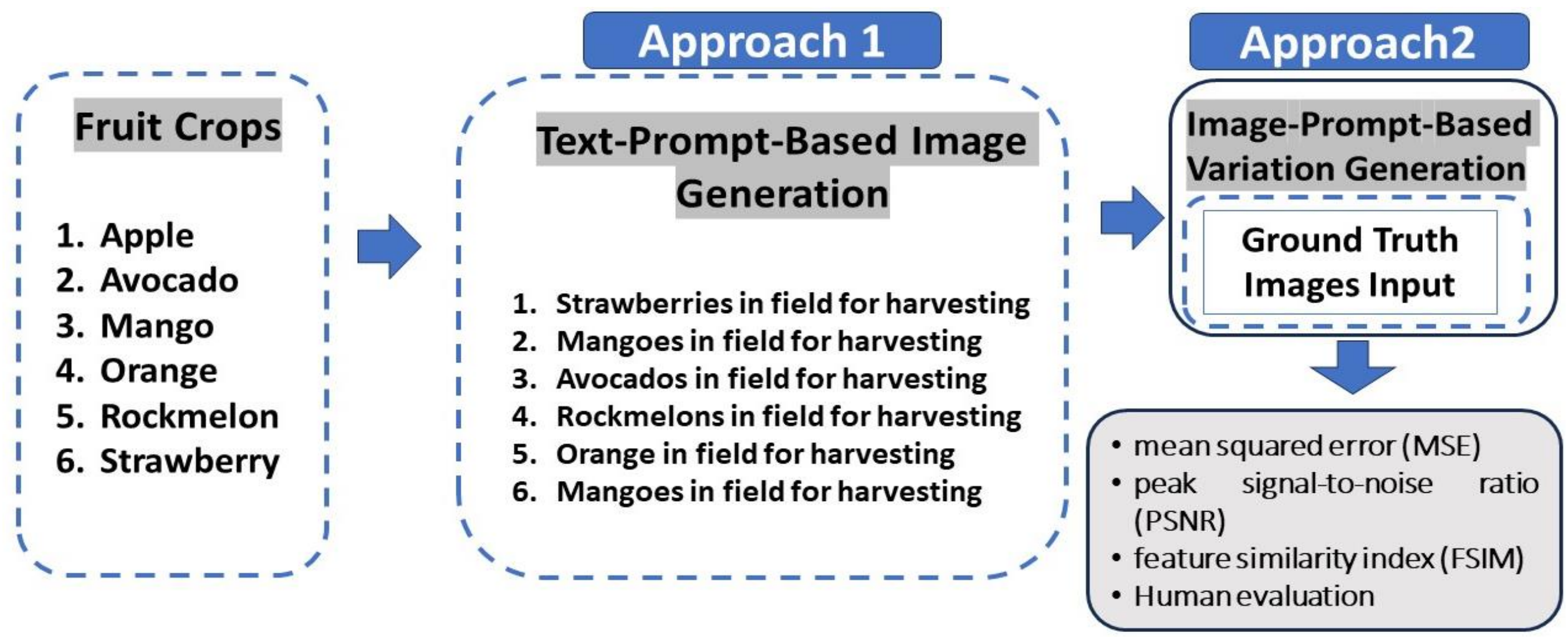

**Figure 2: Flow diagram depicting the Two-Step process utilized to generate agricultural image datasets using by the Generative AI Model DALL.E: The first step involves synthesizing images from textual prompts without any visual input, and the second step generates variations using a ground truth image as a reference.**

1. The dataset encompassed a variety of fruit crops, including strawberries, mangoes, apples, avocados, rockmelons, and oranges. These fruits were carefully selected for their distinctive morphological, textural, and color characteristics, as well as their diverse backgrounds.

The original ground truth images for both datasets were obtained from "A Survey of Public Datasets for Computer Vision Tasks in Precision Agriculture" by Lu and Young (Lu and Young 2020). Six representative images from the fruit crops dataset were randomly selected.

Following this, in the initial image generation step, input text prompts were crafted by carefully examining the original images as depicted in Figure 2. These prompts were then used to generate the category of images. For the second approach, we directly used the ground truth images as input to the DALL.E model to create variations. All the necessary permissions or licenses to collect strawberries, mangoes, apples, avocados, rockmelons, and oranges for this study.

After a processing step, images were generated for both categories. The generated images from both approaches were compared against their respective ground truth images. We evaluated the resulting visuals using key metrics: Peak Signal-to-Noise Ratio (PSNR) for image clarity and pixel accuracy, and Feature Similarity Index (FSIM) for structural similarity. Additionally, human assessments were conducted to confirm their realism.

**DALL.E Image Generation Model**

In this study, DALL·E 2 (OpenAI, California, USA) image generation model was used, which utilizes hierarchical text-conditional image generation to produce images based on textual descriptions (Ding et al. 2022). The hierarchical text-conditional image generation involves a (contrastive model) CLIP image embedding from a given text caption, taking advantage of CLIP's ability to learn robust image representations that encompass both the subject matter and stylistic elements (Conde and Turgutlu 2021). The second stage involves a decoder that creates an image based on this embedding. This method is designed to enhance the variations in the generated images while maintaining their photorealism and relevance to the caption (Ramesh et al. 2022). Additionally, it allows for the generation of image variations that retain the core semantics and style, altering only the incidental details not captured in the image representation. The DALL.E model leverages diffusion models in the decoding phase to discover effective techniques for creating high-quality images. These images can be finely tuned based on textual directions, eliminating the necessity for the model to undergo specialized pre-training for distinct image editing operations.

The model consists of three stage processes: encoder, prior and decoder. The model takes a textual input which is then encoded into a Compact Language-Image Pretrained (CLIP) text embedding based on a neural network trained on hundreds of millions of tax-image pairs. Dimensionality of the resulting CLIP text embedding is then reduced using Principal Component Analysis (PCA) before the results are provided the prior stage. In the prior stage, a Transformer model with an attention mechanism transforms the CLIP text embedding into an image embedding. Following the prior stage, the image embedding goes through the decoder stage, also known as the unCLIP phase, in which a diffusion model based on Generative Adversarial Network (GAN) is used to convert it into an image. The output is subsequently generated through two Convolutional Neural Networks (CNNs) for upscaling: first from 64x64 resolution to 256x256, and then to a final resolution of 1024x1024. The model utilizes semantic components, handling inpainting tasks, and altering images based on subtle changes in the contextual understanding of the input text to produce the output.

### Text-to-image generation

In this study, text prompts displayed in Figure 2 were created to generate images across the six specified categories of fruit crops. These text prompts were carefully designed to ensure that the synthetic images conveyed significant information, closely representing the real images. Initially, a manual analysis of randomly selected ground truth (actual) images was conducted for six fruit crop environments. Text prompts were then crafted based on the visual characteristics of the ground truth images, with input text ranging from a minimum of 4 words to a maximum of 10 words, as illustrated in Figure 2. For all fruit crops, the input text prompts were uniform, describing the "*name of the fruit* in the field for harvesting," where "in the field for harvesting" was a common phrase reflecting the harvesting condition, making a 5-word input text prompt used for each category.

### Image-to-image (variations) generation

In this approach, actual images (ground truth images) representing the specific fruit crop categories were provided to the model as input image prompts as shown in Figure 3. The model was then activated to generate four variations of the given input image upon receiving the command "Generate Variations." An illustration of this approach to image generation is depicted in Figure 3.

## Analysis of the generated images

In this research, the generated images were analyzed to assess their fidelity and realism. Evaluation metrics as PSNR and FSIM were employed to quantify the similarity between AI-generated and ground-truth images. Additionally, human evaluations by 15 scholars from Washington State University, Irrigated Agriculture Research and Extension Center (IAREC) provided subjective insights into the realistic portrayal of these images. Figure 4 depicts the image analysis procedure used. All generated image datasets, obtained through the text-to-image and image-to-image approaches discussed above, underwent a standardized preprocessing procedure. Initially, the images were converted to grayscale and resized to a resolution of 256 by 256 pixels for subsequent pixel-level analysis. For the statistical comparison of the generated images with the respective ground truth images, the images were resized and converted to grayscale as shown in Figure 4.

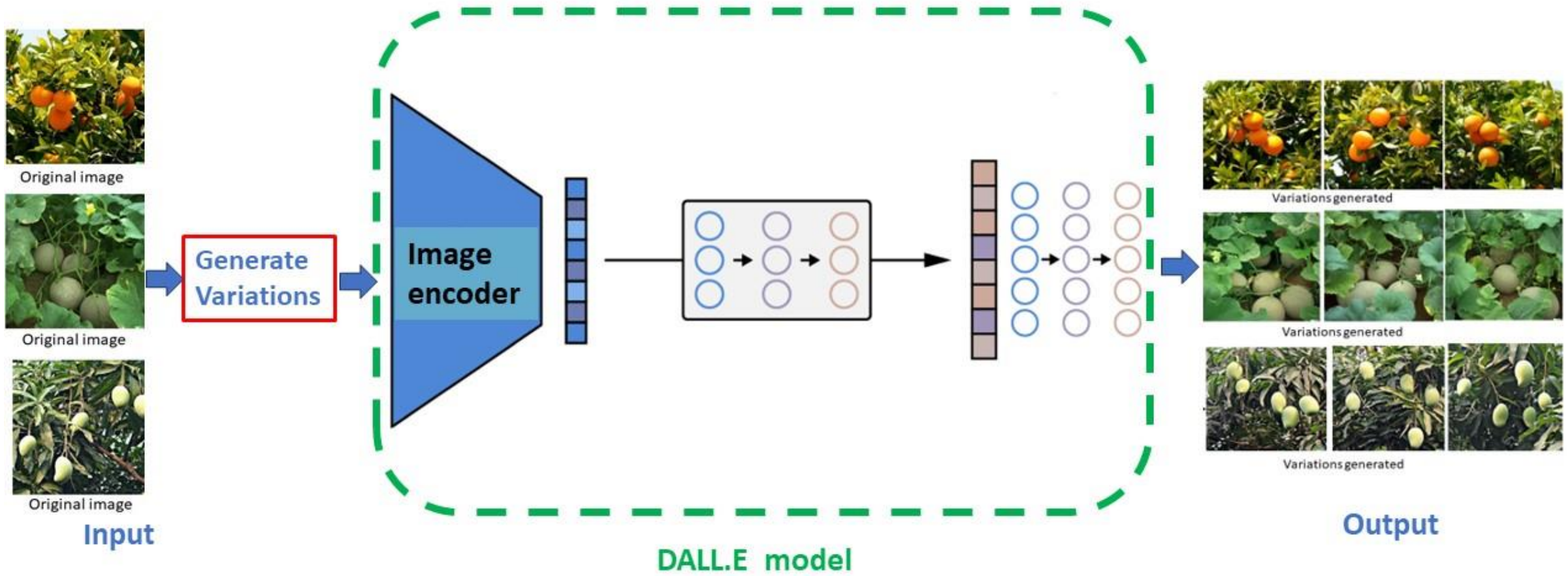

**Figure 3: An example showing image-to-image variation generation using the DALL.E model**

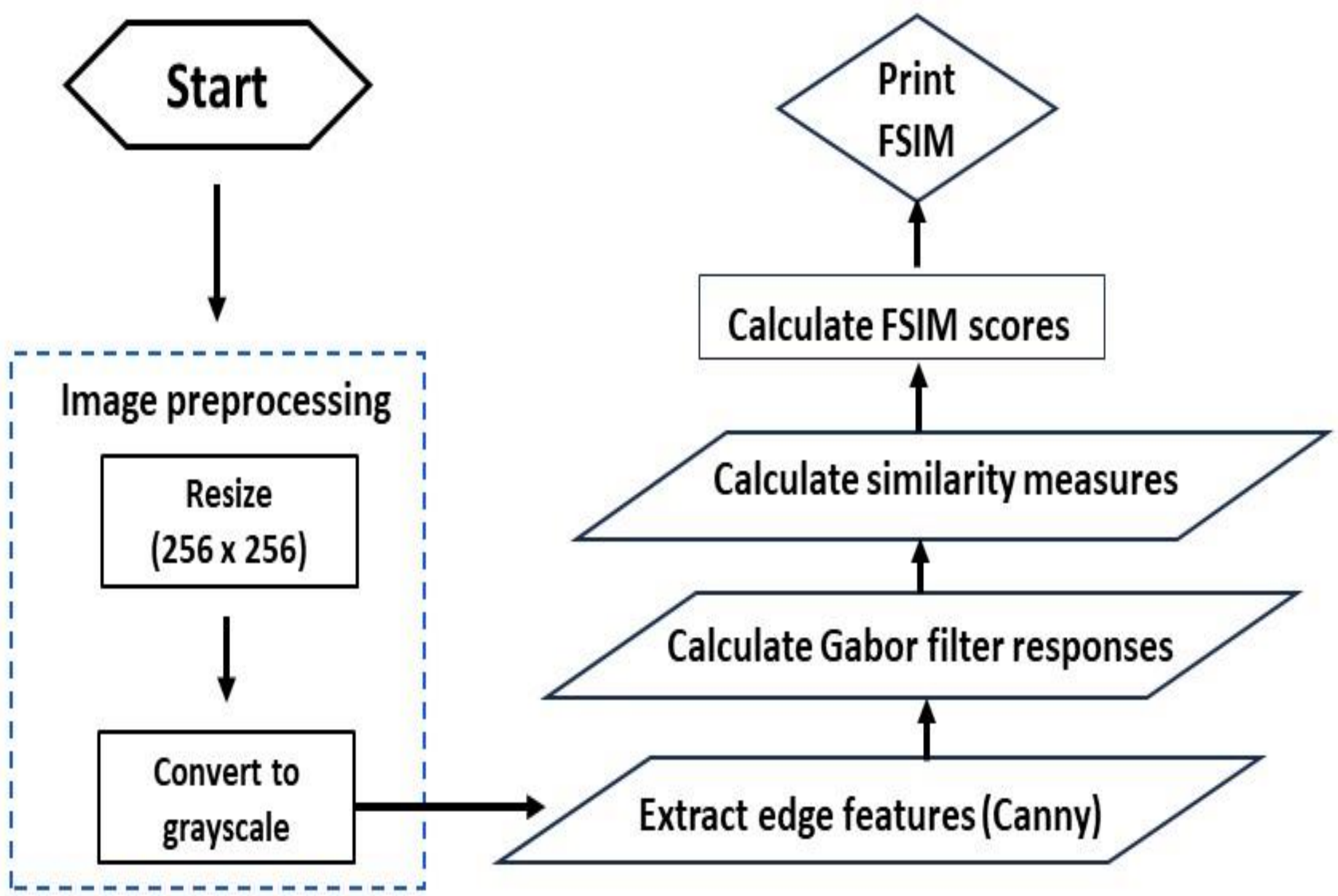

**Figure 4: Block diagram showing the process of calculating feature similarity index (FSIM)**

**Evaluation Measures**

In this study, the images generated in both Step 1 (text-prompt-based image generation) and Step 2 (image-prompt-based image variation generation) were compared against the ground truth images using two standard metrics as follows.

**Peak Signal to Noise Ratio (PSNR):** PSNR served as a metric for assessing image quality by evaluating the ratio of the maximum potential power of the signal (represented by the original image) to the power of disruptive noise (capturing the disparities between the original and the AI-generated image) as given by Equation 1. Mean Squared Error (MSE) estimated using Equation 2 was used to calculate this ratio. A higher PSNR typically indicates that the generated image is closer in quality to the original image and has minimal distortion.

$$PSNR = 10 \log_{10}\left(\frac{MAX_I^2}{MSE}\right)$$

***Equation 1***

Where, $MAX_I$ denotes the maximum possible pixel value in the image and Mean Squared Error (MSE) is computed as the average of the squared differences between corresponding pixels in the two images. In simple words, MSE helps in understanding how much the generated image (G) deviates from the original image (O) on a pixel-by-pixel basis.

$$MSE = (\frac{1}{N})\sum_{i=1}^{N}(O-G)^2$$

***Equation 2***

**Feature Similarity Index (FSIM):** FSIM assesses the similarity between the AI-generated images and the original/actual images based on their features. It evaluates both basic and intricate image features, providing a thorough measure of similarity. FSIM considers aspects like structure, luminance, and contrast of the images. Although the exact calculation of FSIM (Equation 3) involves complex comparisons at multiple scales, the key idea is that it measures how closely the features of the generated image match those of the original image (Zhang et al. 2011). In this analysis, the images were preprocessed by being resized to 256 by 256 pixels and then converted into grayscale images. Canny edge features were then extracted from the grayscale images to calculate the Gabor filter responses. These responses were used to calculate the similarity measures and evaluate the feature similarity score, as shown in Figure 4.

$$FSIM = \prod_{k=1}^{K}\left(\frac{l_k.c_k.s_k}{L_k.C_k.S_k}\right)^{\alpha_k}$$

***Equation 3***

Where, $l_k$ , $c_k$ , and $s_k$ are the local similarity, contrast, and structure measurements at scale k, respectively, and $\alpha_k$ is the weight assigned to each scale.

**Human Evaluation:** A group of 15 agricultural scholars including graduate students and professors from Biological Systems Engineering and Horticulture departments at Washington State University participated in an unbiased survey to evaluate the realism of images generated by the DALL·E model. To ensure independence and minimize bias, the same set of images used for PSNR and FSIM analysis (8 images each for ground truth, text-generated set and image-variation-generated set) was provided to the participants. The participants were unaware of whether the individual images were generated using AI or were acquired in the field using a camera. Each participant was given only the name of the crop environment with no additional information. They used a 5-point likelihood scale to rate the realism, ranging from 'Not at all realistic' (1) to 'Extremely realistic' (5).

## Results and Discussion

The result images generated in this study by the DALL·E model from textual inputs are presented in Figure 5. The model accurately depicted strawberries in the field condition with plastic mulch (Figure 5a), mangoes on tree branches (Figure 5b), mature apples in a tree (Figure 5c), avocados in tree canopies with foliage (Figure 5d), a rockmelon with its characteristic netted skin (Figure 5e), and oranges on tree section realistically as shown in Figure 5f. Each image effectively illustrated the distinct morphological features of the fruits and their environments, showcasing the model's ability to create detailed and contextually precise visual representations from text descriptions. Variations of these fruit crops, based on ground truth images, were also generated by the model. These variations were subtly different yet retained the essence of the original images. For strawberries, variations in the ground cover were shown; avocados were depicted hanging in different positions; apples were presented in various cluster formations; and the oranges, rockmelons, and mangoes were characterized by their vibrant colors, unique textures, and distinct shapes.

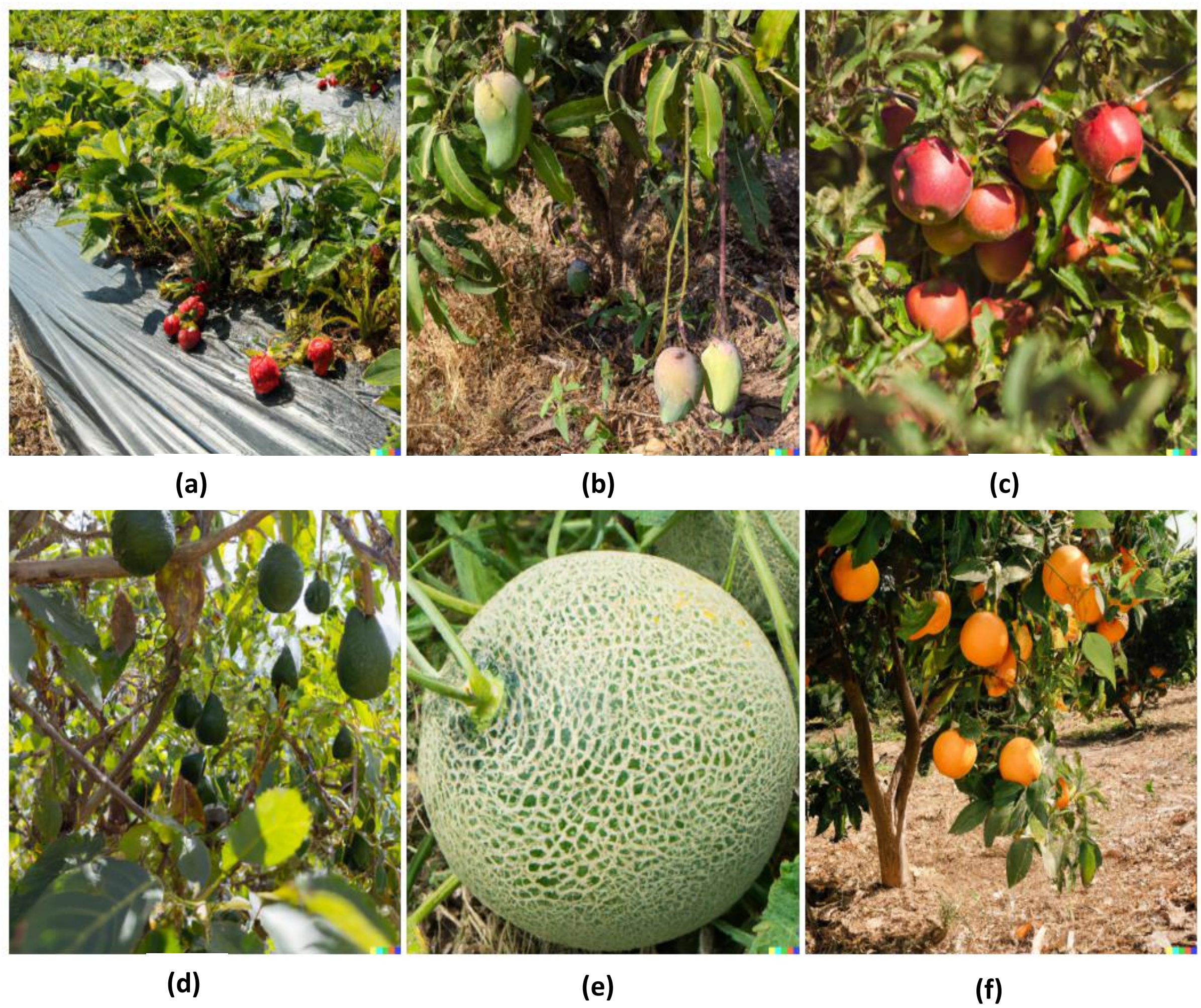

**Figure 5: Fruit crop images generated using text-to-image generation approach using the DALL.E model; (a) strawberries; (b) mangoes; (c) apples; (d) avocados; (e) rockmelon; and (f) oranges.**

Likewise, the image generated by the DALL.E model using the ground truth image as an input to generate image-variations are depicted in Figure 6. Each of the six fruit types including strawberries (Figure 6a), avocados (Figure 6b), apples (Figure 6c), mangoes (Figure 6d), rockmelons (Figure 6e), and oranges (Figure 6f) were characterized by subtle yet realistic modifications, maintaining the essence of the original, ground truth images.

## Quantitative Similarity Measures

For the text-generated images, as shown in Figure 7a the PSNR values ranged from a low of 8.3 for rockmelons to a high of 10.6 for avocados, indicating a variation in the model's ability to replicate image quality. In contrast, the PSNR for image-generated (variation) images was 14.6 for mangoes, suggesting a potential for superior representation of the reality of fruit crop environments. However, the lowest PSNR in this category was 8.8 for strawberries, highlighting a potential weakness in representing reality in a wider range of agricultural environments.

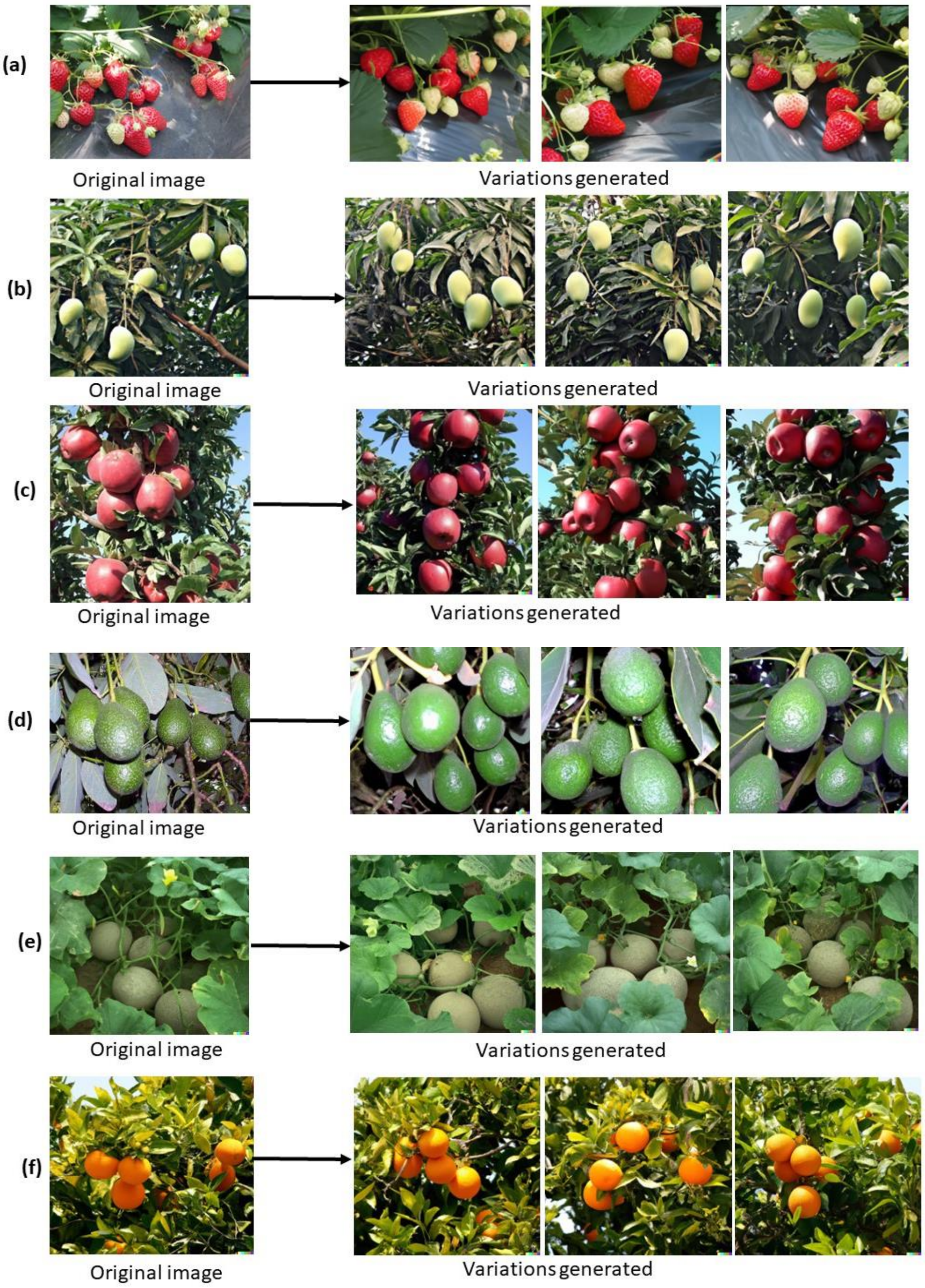

**Figure 6: Three variations (right) of fruit crop images generated by DALL·E 2 model using original images (left) as an input; (a) strawberries; (b) mangoes; (c) apples; (d) avocados; (e) rockmelon; and (f) oranges.**

Likewise, as shown in Figure 7b, the FSIM index revealed that image variation generated images achieved slightly higher similarity with the ground truth images compared to text-to-image generated images. The observation that image variation generation underperforms in comparison to text-to-image generation, as measured by PSNR and FSIM, may initially seem counterintuitive. However, this outcome can be attributed to the inherent differences in the model's approach to generating images from textual versus image prompts. Text-to-image generation relies on the model's understanding and interpretation of textual descriptions to create an image from scratch, potentially allowing the model to "idealize" the output, closely matching key features described in the text while maintaining overall coherence and fidelity. In contrast, image variation generation starts with an existing image and attempts to introduce variations within the constraints of the original image's context. This process may inherently limit the extent to which the model can optimize for clarity and structural similarity, as it must balance between preserving the original image's integrity and introducing meaningful variations. As a result, the variations might introduce or exacerbate minor discrepancies in texture or structural details, which could explain the lower PSNR and FSIM scores. This suggests a trade-off in the model's performance between generating novel images from textual descriptions and modifying existing images to create variations, highlighting the challenges in achieving both high fidelity and meaningful diversity in generated images.

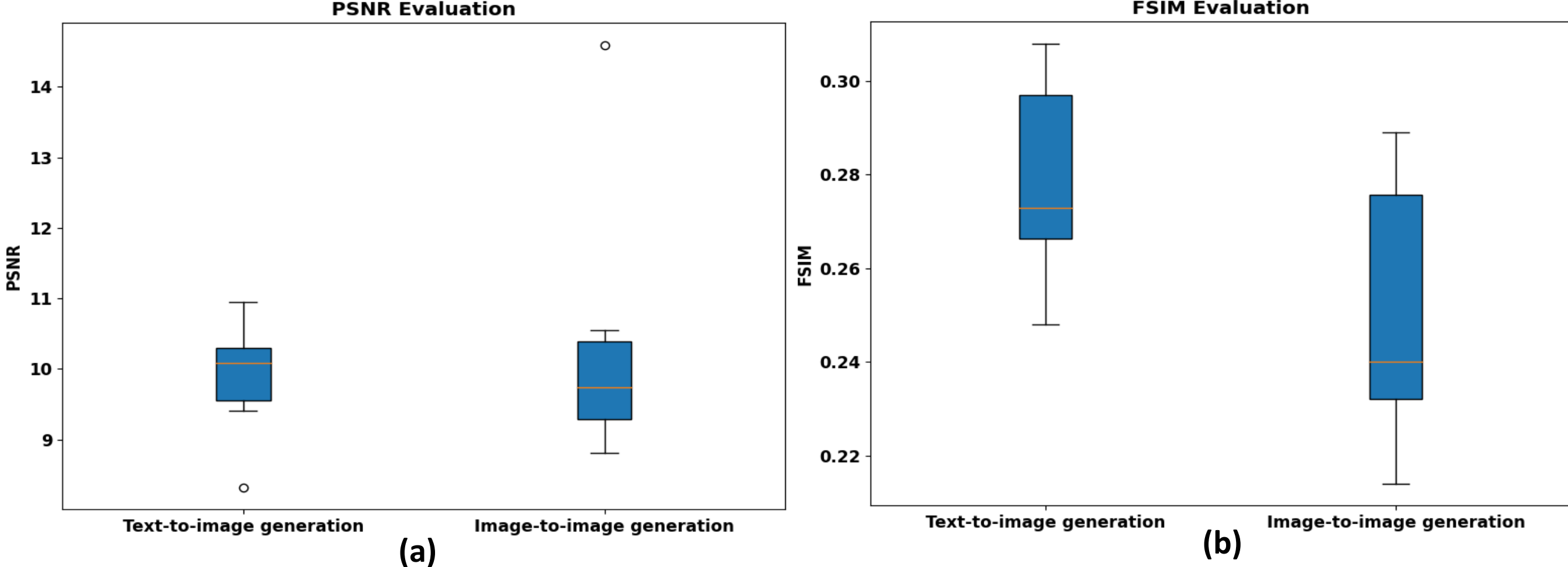

**Figure 7: Box plots illustrating the distribution of PSNR and FSIM for all fruit crops tested; a) comparing text-to-image; b) image-to-image generation methods in agricultural AI applications.**

These results suggest a generally lower performance in maintaining feature similarity compared to the original images, particularly in the case of avocados (0.2) when compared to text-generated images. While both text and image-prompt approaches displayed strengths in certain aspects, there were notable variations in performance across different fruit types and metrics. This analysis indicates that the model's effectiveness in generating accurate and realistic images in diverse agriculture environment is promising but is dependent on crop types and cropping environments.

## Human Evaluation Results

Results of the human assessment of the AI-generated and original images for all six fruit crops are depicted in Figure 8. In the text-to-image category, apples consistently received high ratings, indicating a strong capability of the AI model to interpret textual prompts and generate realistic visual representations. On the other hand, Avocados recorded lower ratings, suggesting challenges in capturing their unique textures, colors and/or other features through text descriptions alone. In generating the image-to-image variations, Mangoes and Rockmelons received notably high ratings, showcasing the model's proficiency in creating realistic variations from existing images. The lower ratings for Strawberries in this category might reflect difficulties in maintaining the fruit's distinct characteristics in generating variations. Ground truth images, as expected, generally received the highest ratings across all categories, affirming their authenticity, which also indicated that there is a huge room for improvement in AI modeling to replicate complexities in the plant canopies and agricultural fields.

Despite those challenges, it is noted that there were instances where text-based or image-based AI generations outperformed the original images in specific fruit crops. For example, image-to-image

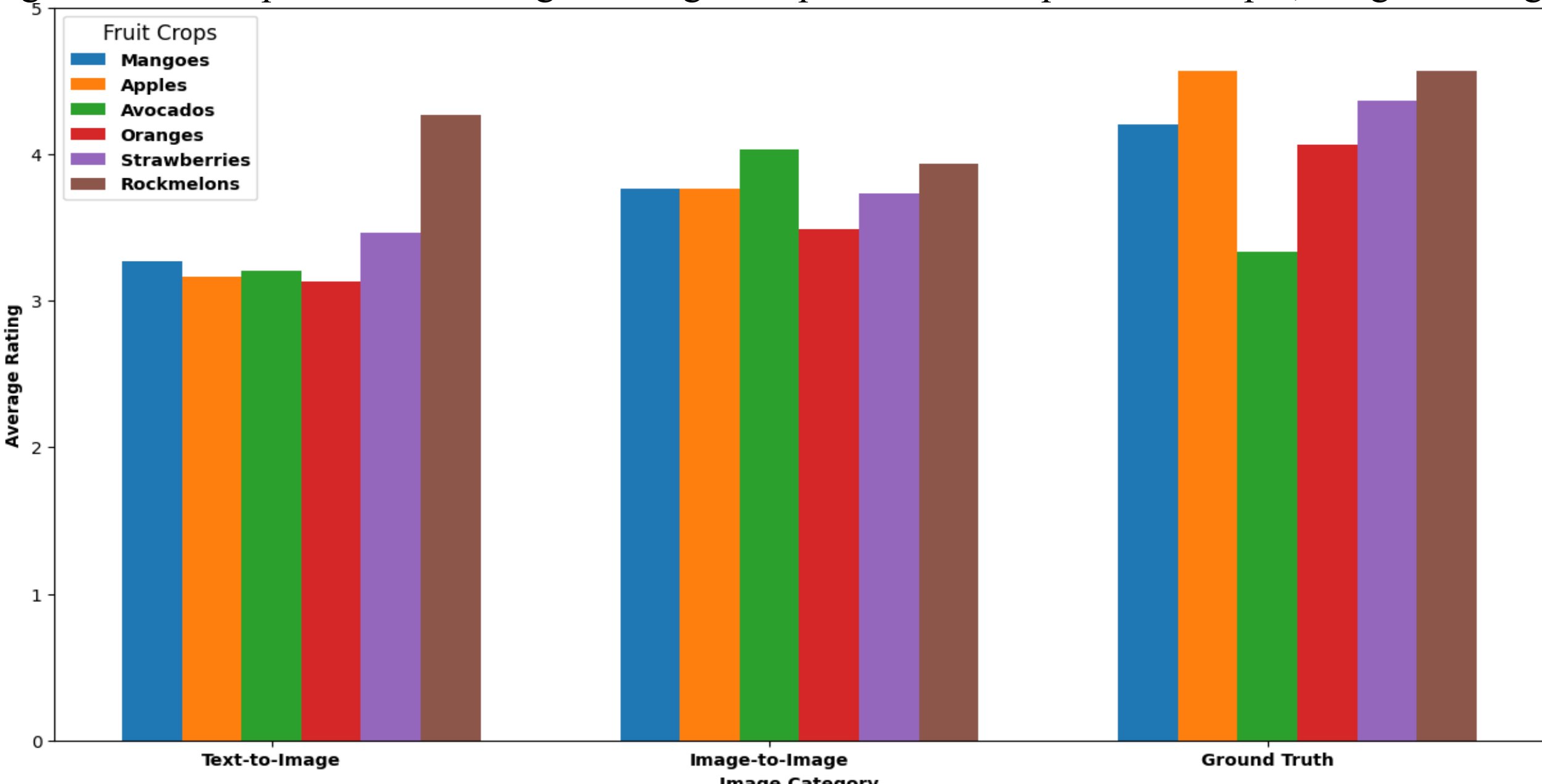

**Figure 8: Bar chart illustrating average human evaluation ratings for Text-to-Image, Image-to-Image variations, and Ground Truth across six different fruit crops images in this survey of image generation process using Generative AI**

variations of Mangoes and Rockmelons occasionally surpassed ground truth ratings. This could be attributed to the AI's ability to enhance certain visual aspects, such as color vibrancy or clarity, making them more appealing than the actual photographs to human observers. The success in these instances shows the potential of AI-based image generation to not only replicate but potentially improve upon real-world images of agricultural fields.

Figure 9 presents a heatmap detailing the MSE, PSNR, and FSIM for both Text-to-Image and Image-to-Image (variation) generated outputs. For fruit crops, text-to-image generation yielded PSNR values up to 10.95 (for avocados), indicating a commendable image quality. However, image-based variations surpassed this, with rockmelons achieving a higher PSNR of 14.592, suggesting a closer resemblance to actual images. The Feature Similarity Index (FSIM) echoed this trend; while text-generated images like mangoes scored 0.308, indicating satisfactory structural similarity, image-based generation scored marginally lower at 0.287 for the same fruit.

These results indicate that while text-based generation showed notable proficiency, image-based variations generally offered enhanced clarity and fidelity. The best results was achieved for rockmelons image-based generation, whereas the lowest was in text-based generation for avocados. This pattern suggests that while AI applications like DALL.E in this study can generate reasonably accurate representations from textual descriptions, providing image prompts leads to more precise and realistic visual outputs, especially in complex agricultural scenarios.

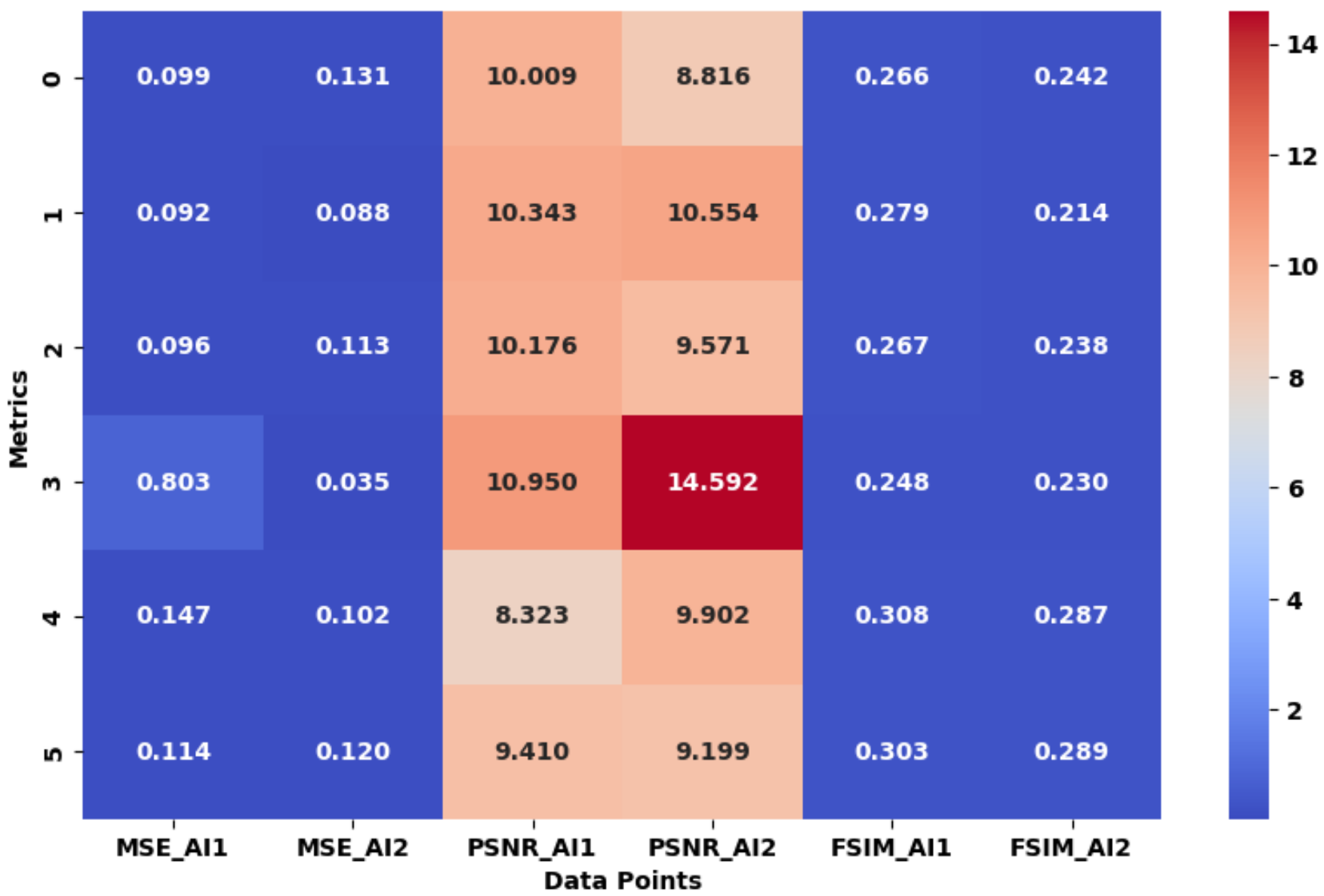

**Figure 9: Heatmap for the DALL.E Model Performance: Showcases a detailed comparison of MSE, PSNR, and FSIM metrics across the six fruit crops**

The image-to-image generation method consistently outperformed the text-to-image approach. Notably, rockmelons generated through image-to-image variations exhibited high clarity and detail, as evidenced by superior PSNR score of 14.6. This was indicative of the AI's proficiency in producing clear and detailed images, essential for tasks like crop growth monitoring and yield estimation. Mango crop images, generated through both methods, showcased high structural similarity with the actual images achieving an FSIM score of 0.31 for text-to-image generation method and 0.29 in image-to-image generation method. These results also highlight the AI's capability to maintain structural integrity, crucial for shape-based agricultural tasks such as on fruit crops results.

These results demonstrated that the DALL.E model can be used to generate large image datasets for agricultural applications with good accuracy and structural integrity. Such AI-generated images can significantly simplify the data generation process, reducing time and costs associated with traditional methods that rely on advanced sensors and intensive field data collection.

The findings suggest that DALL.E 2's capabilities in image generation hold potential for advancing machine vision and robotic operations in agriculture, contributing to the development of more efficient and accurate AI-driven agricultural systems and accelerate their adoption. Previous studies for image generation in agricultural environments typically depended on labor-intensive and expensive field data collection, often hindering efficiency. However, in this study, we showed an efficient workflow of creating agricultural images using AI that could potentially avoid the reliance on labor-intensive and costly field data collection methods in the near future. Our evaluation of the feature similarity between AI-generated images and real sensor-captured images of crop environments not only validates the practical utility of this technology but also opens up new possibilities for its application in precision agriculture. This shift towards AI-generated imagery could potentially revolutionize the way agricultural studies are conducted, offering a more cost-effective, rapid, and versatile method of data collection.

## Conclusion and Future Prospects

In modern agriculture, the need for comprehensive image datasets is paramount, especially given the limitations of traditional data collection methods, which are often labor-intensive and time-consuming. Synthetic image generation emerges as a compelling solution, addressing these challenges by creating realistic and diverse datasets efficiently. The utilization of AI-based methods, particularly the DALL.E model developed by OpenAI, exemplifies this approach. Functioning similarly to its counterpart ChatGPT, DALL.E is trained on a vast array of images and textual data, enabling it to generate accurate and diverse images from textual descriptions and existing images. The DALL.E model's potential in agricultural applications is, therefore, substantial. It offers innovative solutions for critical tasks such as fruit quality assessment, automated harvesting, and crop yield estimation. By generating realistic images of various fruit crops, DALL.E aids in the development of smart farming techniques. For instance, the model's ability to create images of fruits in different growth stages can help in training AI models for precise fruit detection, thus improving crop monitoring and harvesting strategies. Additionally, there could potentially come a day soon when we will need no sensors and tedious image collection procedure to perform RGB data acquisition due to the recent advancement of LLMs as DALL.E. This study conducted a detailed evaluation of the DALL.E model's efficacy in generating agriculturally relevant images, focusing on its ability to replicate and enhance real-world field conditions through synthetic imagery.

Based on the results, the following specific conclusions can be made from this study for the AI based image generation using DALL.E in six diverse fruit crops in agriculture:

- Image-to-image generation methods resulted in a 5.78% increase in average PSNR, indicating improved image clarity and quality over text-to-image generation.
- However, there was a decrease of 10.23% in average FSIM for image-to-image generation, suggesting a reduction in structural and textural similarity to the original images compared to text-to-image generation.

This study underscores the potential transformative impact of integrating advanced AI model DALL.E into advancing agricultural technologies and solutions. The successful application of this model in generating realistic images for various agricultural scenarios opens new opportunities for enhancing agricultural efficiency and improving crop yield and quality in the coming days. By leveraging the capabilities of the generative AI model DALL.E, which is a Large Language Models (LLMs), the agricultural sector could see a significant shift in how data is gathered and analyzed. The traditional reliance on sensors and manual data collection processes, often cumbersome and time-intensive, could be greatly reduced or even be completely replaced in the future. Instead, AI-generated images, as demonstrated in this study, could provide a more efficient and scalable alternative. The ability of models like DALL.E to create accurate depictions of diverse agricultural environments from different fruits in diverse realistic backgrounds offers new potential for smart and precision agricultural practices. In the future, tasks like yield estimation, disease detection, and crop health monitoring could be conducted using datasets generated entirely by AI, streamlining the process and increasing its accuracy and adaptability.

Looking ahead, the trajectory of generative AI in agriculture is set to expand with models like DALL.E already laying the groundwork for synthetic dataset creation. While this study focused on leveraging DALL.E for text-to-image generation, future advancements could see the inclusion of text-to-video generator models like SORA by OpenAI, potentially offering even more powerful and useful applications. The recent ongoing development of AI Model such as SORA text to video generator could be instrumental in generating dynamic visual content that captures the complexities of agricultural environments across time, complementing DALL.E's capabilities.

## Acknowledgements and Funding

The authors wish to express their profound gratitude to the following scholars for their invaluable contribution in assessing the realism of the images evaluated in this study: Dr. Sindhuja Sankaran, Dr. Joan Wu, Dr. Markus Keller, Dr. Safal Kshetri, Dr. Salik Khanal, Bernardita Veronica , Achyut Paudel, Datta Bhalekar, Shafik Kiraga, Alexander You, Karisma Yumnam , Elda Yitbarek Bizuayene, Atif Bilal Asad, Dr. Chenchen Kang, Priyanka Upadhayaya, Martin Churuvija and Syed Usama Bin Sabir.

This work was supported in part by the National Science Foundation (NSF) and the United States Department of Agriculture (USDA), National Institute of Food and Agriculture (NIFA), through the "Artificial Intelligence (AI) Institute for Agriculture" program under Award Numbers AWD003473 and AWD004595, and USDA-NIFA Accession Number 1029004 for the project titled "Robotic Blossom Thinning with Soft Manipulators." Additional support was provided through USDANIFA Grant Number 2024-67022-41788, Accession Number 1031712, under the project "ExPanding UCF AI Research To Novel Agricultural EngineeRing Applications (PARTNER)."

## Competing interests

The authors declare no competing interests.

## Additional information

Correspondence and requests for materials should be addressed to R.S. and/or M.K.